\documentclass[10pt,twocolumn,letterpaper]{article}

\usepackage{cvpr}
\usepackage{times}
\usepackage{epsfig}
\usepackage{graphicx}
\usepackage{amsmath}
\usepackage{amssymb}
\usepackage{paralist}

\usepackage[linesnumbered,ruled]{algorithm2e}
\usepackage[table]{xcolor}
\usepackage{booktabs}
\usepackage[pagebackref=true,breaklinks=true,letterpaper=true,colorlinks,bookmarks=false]{hyperref}

\cvprfinalcopy 

\def\cvprPaperID{0124} 

\ifcvprfinal\fi
\begin{document}

\title{ProNet: Learning to Propose Object-specific Boxes\\ for Cascaded Neural Networks}

\author{Chen Sun$^{1,2}$\quad Manohar Paluri$^2$\quad Ronan Collobert$^2$
       \quad Ram Nevatia$^1$ \quad Lubomir Bourdev$^3$
       \\
       \begin{tabular}{ccc}
       $^1$ USC & $^2$ Facebook AI Research & $^3$ UC Berkeley\\
        {\tt\small\{chensun, nevatia\}@usc.edu} & {\tt\small\{mano, locronan\}@fb.com} & {\tt\small lubomir.bourdev@gmail.com}
       \end{tabular}
       }

\maketitle

\begin{abstract}
This paper aims to classify and locate objects accurately and efficiently, without using bounding box annotations. It is challenging as objects in the wild could appear at arbitrary locations and in different scales. 
In this paper, we propose a novel classification architecture ProNet based on convolutional neural networks. 
It uses computationally efficient neural networks to propose image regions that are likely to contain objects, and applies more powerful but slower networks on the proposed regions. The basic building block is a multi-scale fully-convolutional network which assigns object confidence scores to boxes at different locations and scales. We show that such networks can be trained effectively using image-level annotations, and can be connected into cascades or trees for efficient object classification. ProNet outperforms previous state-of-the-art significantly on PASCAL VOC 2012 and MS COCO datasets for object classification and point-based localization.
\end{abstract}



\section{Introduction}

We address the problem of object classification and localization in natural images. As objects could be small and appear at arbitrary locations, several frameworks~\cite{DBLP:conf/cvpr/OquabBLS14,NUS-PSL} rely on bounding boxes to train object-centric classifiers, and apply the classifiers by searching over different locations of the images. However, the annotation process for object bounding boxes is usually resource intensive and difficult to scale up. In light of this, we aim to simultaneously classify and locate objects given only \textit{image-level annotations} for training.

\begin{figure}[t]
  \centering
    \includegraphics[width=\linewidth]{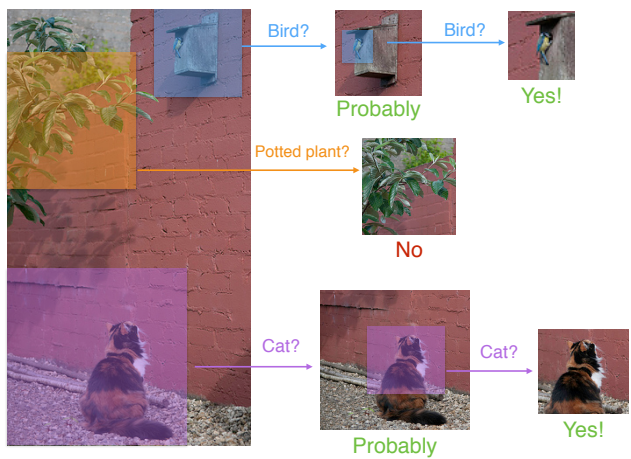}
  \caption{We approach object classification problem by zooming onto promising image boxes. No object-level annotations are needed during training.}
  \label{fig:teaser}
\end{figure}

To cope with the lack of object-level annotations, several methods~\cite{DBLP:journals/corr/ChatfieldSVZ14,DBLP:journals/corr/GongWGL14,Simonyan14c} extract feature activations from convolutional neural networks (CNN) by scanning over different image regions. They then aggregate the extracted features into image-level representations for classification purpose. Under this scheme, regions that belong to the background are considered as important as regions that contain objects. Such global approaches tend to be sensitive to background, and cannot be used directly for localization.

We choose to use the fully-convolutional network (FCN) architecture~\cite{DBLP:journals/corr/LongSD14,Oquab_2015_CVPR,pinheiro:2015a,DBLP:journals/corr/SermanetEZMFL13,DBLP:journals/corr/PathakSLD14} for simultaneous object classification and localization. It replaces the fully-connected layers of a standard CNN (e.g. AlexNet~\cite{NIPS2012_4824}) with convolutional layers. This enables an FCN to take images of arbitrary sizes, and generate classification score maps efficiently. Each element in a score map corresponds to a rectangular box (\textit{receptive field}) in the original image. The score maps can then be used for classification and localization.

The sampling strides and box sizes are determined by the FCN's network architecture. As box sizes are fixed, FCN might face difficulty dealing with objects of different scales. We address this problem by using a multi-stream multi-scale architecture. All streams share the same parameters, but take input images of different scales. To train the multi-scale FCN without object-level annotations, we generate image-level scores by pooling the score maps over multiple-scales, and compute the losses with image-level labels for back-propagation.

Once a multi-scale FCN is trained, it can be used for classification and localization directly. From another perspective, it also proposes a set of \textit{promising boxes} that are likely to contain objects. We can then build a \textit{cascade} architecture by zooming onto those promising boxes, and train new classifiers to verify them. The cascade allows the system to balance accuracy and speed: each stage filters out parts of image regions that are unlikely to contain objects. We name this \textit{propose} and \textit{zoom} pipeline as \textbf{ProNet}. Figure~\ref{fig:teaser} provides the high-level intuition behind ProNet: three boxes are proposed for \textit{bird}, \textit{potted plant} and \textit{cat} categories. The boxes are cropped out and verified further, until a certain decision is made.

To train the later classifiers in ProNet, we sample \textit{hard negatives} based on image-level labels. For positives, as no object-level annotations are available, it is impossible to tell objects from background. To avoid over-fitting, we randomly sample positive boxes above a relative low threshold. Different positive boxes from the same image can be sampled at different iterations of the stochastic gradient descent training process. At test time, only a small subset of boxes (10 to 20 per image) with highest object confidence scores are fed to the later classifiers. This allows us to utilize CNNs that have stronger representation power with little computational overhead.

ProNet is highly configurable: for example, one could set a list of important object categories, and only verify the proposed boxes for those categories. Moreover, apart from a traditional chain-structured cascade, we show that it is also possible to build tree-structured cascades, where each branch handles categories from a particular domain (\eg set of \textit{vehicles} or \textit{animals}). 

In summary, our paper makes the following contributions:
\begin{compactitem}
\item We propose ProNet, a cascaded neural network framework that zooms onto promising object-specific boxes for efficient object classification and localization.
\item We introduce strategies to train ProNet with image-level annotations effectively; and demonstrate the implementations of chain- and tree-structured cascades.
\item We show that ProNet outperforms previous state-of-the-art significantly on the object classification and point-based localization tasks of the PASCAL VOC 2012 dataset and the recently released MS COCO dataset.
\end{compactitem}

\section{Related Work}
Object classification is a fundamental problem in Computer Vision. Earlier work~\cite{DBLP:conf/iccv/GraumanD05,DBLP:journals/cviu/Fei-FeiFP07,DBLP:journals/ijcv/SanchezPMV13} focused on classification from object-centric images. They usually extract hand-crafted low-level features~\cite{DBLP:journals/ijcv/Lowe04} and aggregate the features into image-level feature vectors~\cite{Perronnin07FV,DBLP:conf/cvpr/WangYYLHG10}. More challenging datasets~\cite{Everingham10,DBLP:journals/corr/LinMBHPRDZ14,ILSVRCarxiv14} have since been collected. They are of larger scale, and contain smaller objects which could be partially occluded.

Recently, deep convolutional neural networks (CNN) have achieved state-of-the-art performance on a wide range of visual recognition tasks, including object classification~\cite{NIPS2012_4824,Simonyan14c,DBLP:journals/corr/SermanetEZMFL13} and detection~\cite{DBLP:journals/corr/GirshickDDM13,DBLP:journals/corr/Girshick15}. Although CNNs require large amount of data for training, it has been shown that they are able to learn representations that generalize to other tasks. Such representations can be adapted to image classification by fine-tuning~\cite{DBLP:conf/cvpr/OquabBLS14}, or extracted as holistic features for classification with linear SVMs~\cite{DBLP:journals/corr/ChatfieldSVZ14}. When used as generic feature extractors, feature aggregation techniques designed for hand-crafted features can also work with CNN embeddings and achieve competitive performance~\cite{DBLP:journals/corr/GongWGL14}.

\begin{figure*}[!ht]
  \centering
    \includegraphics[width=0.95\linewidth]{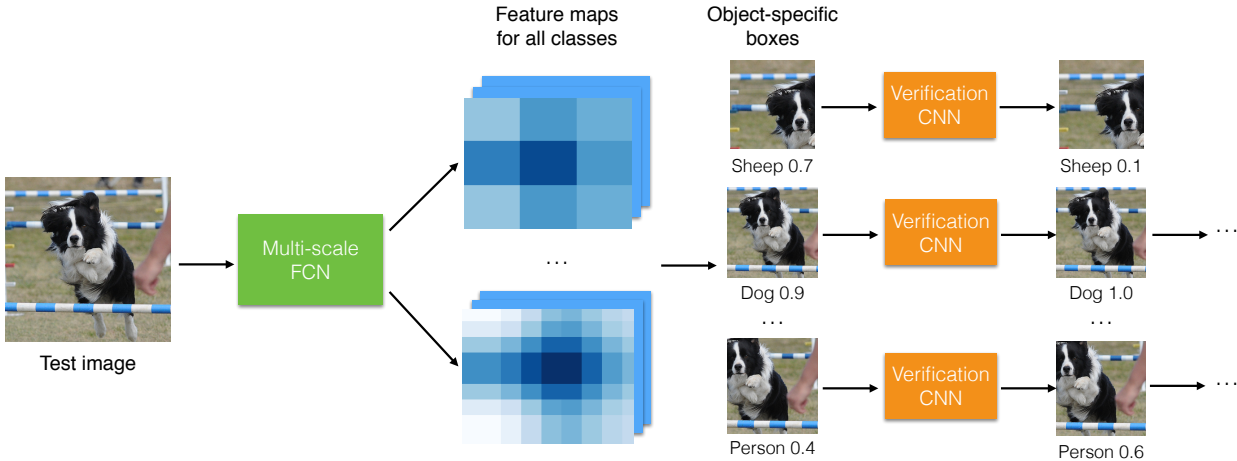}
  \caption{Illustration of the proposed ProNet framework. Given a test image, it first applies a multi-scale fully-convolutional network to select boxes that are likely to contain objects. It then feeds the selected boxes to CNNs trained on harder instances for verification. CNNs at different levels are connected as chains or trees, and trained in a cascade fashion.}
  \label{fig:flow}
\end{figure*}

An alternative approach for object classification is via detection. Among those utilizing bounding box annotations, RCNN~\cite{DBLP:journals/corr/GirshickDDM13} achieves competitive performance by directly representing image boxes with CNN features and learning classifiers on top of the features. Object proposal techniques~\cite{DBLP:journals/ijcv/UijlingsSGS13,DBLP:conf/eccv/ZitnickD14} are used to sample the image patches for classification. A recent framework, fast RCNN~\cite{DBLP:journals/corr/Girshick15}, uses fully-convolutional networks (FCN) to generate box-level features in batch, and is thus more computational efficient.

Object localization with image-level annotations is a weakly-supervised problem. It can be formulated as a multiple instance learning problem, and has been addressed to learn concept detectors from Internet data~\cite{chen2013neil,DBLP:conf/cvpr/DivvalaFG14,wu2015webconcept}. It has also been studied for object detection~\cite{Bilen_2015_CVPR,DBLP:conf/cvpr/CinbisVS14,DBLP:conf/icml/SongGJMHD14,DBLP:conf/eccv/RussakovskyLYF12,zhou2015cnnlocalization} and segmentation~\cite{DBLP:journals/corr/DaiH015,DBLP:journals/corr/PapandreouCMY15,pinheiro:2015a}. For object classification, Wei et al.~\cite{DBLP:journals/corr/WeiXHNDZY14} treat images as bags of patches, where the patches are selected using objectness criteria. They then use max pooling to fine-tune CNNs based on image-level annotations. Oquab et al.~\cite{Oquab_2015_CVPR} follow a similar approach, but make the training process end-to-end by converting CNNs into FCNs. The proposal generation network in ProNet is also based on FCN, but uses a multi-stream architecture and cross-scale LSE pooling to achieve scale-awareness.

Cascaded classifiers~\cite{DBLP:conf/cvpr/ViolaJ01} are a well-studied technique in Computer Vision. Cascades with CNNs have been explored for facial point detection~\cite{DBLP:conf/cvpr/SunWT13}, face detection~\cite{DBLP:conf/cvpr/LiLSBH15} and pose estimation~\cite{DBLP:journals/corr/ToshevS13}. However, such methods require fully annotated training examples. ProNet adopts the cascade philosophy to balance speed and accuracy, but does not require object bounding boxes for training. Since ProNet is a general object classifier, it can also be extended to have tree structure, where each leaf is a domain expert.

\section{ProNet Framework}

ProNet has two basic components: an object-specific box proposal unit, and a verification unit. For each image, for each object category, the box proposal unit generates a list of confidence scores of the presence of the object instances, and the $(x,y)$ coordinates indicating the locations of the objects. ProNet then zooms onto image boxes with higher scores to further verify if they are positive or hard negatives. The verification units can either take all boxes, which forms a chain structure; or a subset of boxes corresponding to certain domains (\eg \textit{animal}), which forms a tree structure. We implement these two units with convolutional neural networks. Figure~\ref{fig:flow} illustrates the overall ProNet framework.


\subsection{Proposal Generation}
The first stage in our framework is to generate object-specific box proposals with CNNs. For an input image $\mathcal{I}$ and object category $c$, we want to learn a proposal scoring function
\begin{align*}
P(\mathcal{I}, c, \mathbf{l}) \in \mathbb{R}
\end{align*}
where $\mathbf{l} = \{x_1, y_1, x_2, y_2\}$ corresponds to the location of a rectangular image region denoted by its top left and bottom right corners.

A typical CNN architecture for image classification task (e.g. AlexNet~\cite{NIPS2012_4824}) involves a hierarchy of convolutional layers and fully connected layers. The convolutional layers operate on local image patches to extract feature representations. For a $W \times H$ color image with 3 channels, the convolutional layers generate a feature map of $D\times W' \times H'$ elements, where $D$ is the output feature dimension. $W'$ and $H'$ correspond to the width and height of the feature map, they are controlled by input image size, as well as the kernel size, sampling step and padding size of the convolutional layers. The fully connected layers serve as classifiers which take fixed-size inputs, thus require the width and height of input images to be fixed. Therefore, one possible way to compute $P(\mathcal{I}, c, \mathbf{l})$ is to enumerate locations and scales in a sliding window fashion or with bounding box proposals, and feed such image regions to CNNs.

We take an alternative approach based on fully convolutional networks (e.g. OverFeat~\cite{DBLP:journals/corr/SermanetEZMFL13}). Fully convolutional networks (FCN) do not contain fully-connected layers. Rather, they use only the convolutional layers, which allows them to process images of arbitrary sizes. The outputs of FCNs are in the form of $C\times W' \times H'$ feature maps, where $C$ is the number of categories. Each element in a feature map corresponds to the activation response for a particular category over a certain region. Such regions are called \textit{receptive fields} for the activations. Compared with region sampling with sliding windows or bounding box proposals, FCNs offer a seamless solution for end-to-end training under the CNN framework, and also naturally allow the sharing of intermediate features over overlapping image regions.

\textbf{Scale adaptation with multi-stream FCNs.} One issue in use of FCNs is that the sizes of receptive fields are typically fixed, while the object scales may vary a lot. We address this problem by using a multi-stream architecture. 

Assume an FCN has been trained with inputs where objects have been resized to the same scale. We expand the network into $K$ streams, where every stream shares the same parameters as the pre-trained one. Given an image $\mathcal{I}$, we scale it to different sizes $\{\mathcal{I}_1, \mathcal{I}_2, ..., \mathcal{I}_K\}$ and feed to the $K$-stream FCN. The output feature map of each stream corresponds to a different scale in the original image.





\textbf{Training with image-level annotations.} When object bounding boxes are available, training FCNs is straight-forward: one could either crop images with the bounding boxes, or use a loss function which operates directly on feature maps and takes the object locations into account. As such supervision is absent, we need to aggregate local responses into global ones so that image-level labels can be used for training. We use the \textit{log-sum-exp} (LSE) pooling function applied by~\cite{pinheiro:2015a} for semantic segmentation:
\begin{equation}
s_{c} = r^{-1}\log\left[\frac{1}{M}\sum_{x,y,k}\exp(r\cdot s_{c,x,y,k})\right]
\end{equation}
where $c$ is the category, $k$ corresponds to the $k$-th stream of FCN, $x,y$ correspond to location in the feature map, $M$ is the total number of such elements and $r$ is a hyper parameter. The function's output is close to average when $r$ is small and maximum when $r$ is large. Setting $r$ larger makes the aggregation focus on a smaller subset of image boxes, and has the potential to handle smaller objects better.

\begin{figure}
  \centering
    \includegraphics[width=1.05\linewidth]{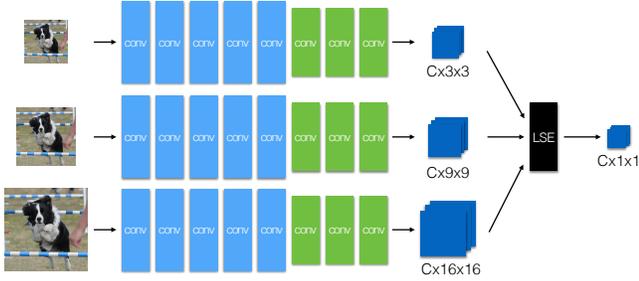}
  \caption{Illustration of 3-stream FCN with LSE pooling.}
  \label{fig:lse}
\end{figure}

LSE pooling function can be implemented as a layer in a neural network. As illustrated in Figure~\ref{fig:lse}, it is connected to the final layers of all $K$-stream FCNs and produces a $C$ dimensional vector for each image. We then compute the loss for each category and back-propagate the error gradients to the earlier layers.

\textbf{Computing proposal scores.} Once the FCNs have been trained, we compute proposal scores $P(\mathcal{I}, c, \mathbf{l})$ from the feature maps. Specifically, for every neuron in the final layer of single-stream FCN, we compute its receptive field and use it as the location $\mathbf{l}$; the corresponding activation of the neuron is used as proposal score. 

Although the exact receptive field may vary due to different padding strategies, we use a simple estimation which has been reported to work well in practice~\cite{DBLP:journals/corr/SermanetEZMFL13}. Denote the sampling stride of a spatial convolutional layer $C_i$ as $d_{C_i}$ and the kernel size of a max pooling layer $M_j$ as $k_{M_j}$, the overall sampling stride $D$ is given by
\begin{equation}
D = \prod_{\mathcal{C}} d_{C_i} \cdot \prod_{\mathcal{M}} k_{M_j}
\end{equation}
where $\mathcal{C}$ is the collection of all convolutional layers and $\mathcal{M}$ is the collection of all max pooling layers.


\textbf{Implementation.} Our $K$-stream FCNs are implemented with Torch. For each stream, we use the CNN-M 2048 architecture proposed in \cite{DBLP:journals/corr/ChatfieldSVZ14}. It has 5 convolutional layers and 3 fully-connected layers. It achieves higher accuracy on ImageNet than AlexNet, while being faster and less memory consuming than very deep CNNs~\cite{Simonyan14c}. We use the model parameters released by the authors, which were pre-trained from ImageNet dataset with 1,000 categories. We convert the model into an FCN by replacing the three fully-connected layers with convolutional layers. The first convolutional layer has 512 input planes, 4096 output planes and kernel size of 6. The second has 4096 input planes, 2048 output planes and kernel size of 1. Since the final layer is task-specific, it is initialized from scratch with 2048 input planes, $|C|$ output planes and kernel size of 1. To adapt the model parameters for object classification on different datasets, we only fine-tune the final two layers and freeze the model parameters from previous layers. The sampling stride of feature maps is 32 pixels, and the  window size is 223 pixels.

We set the number of streams to be 3. During training, all three streams share the same set of parameters. To facilitate training with mini-batches, every image is rescaled to $300\times 300$, $500\times 500$ and $700\times 700$ pixels. As the aspect ratios of images could be different, we rescale the longer edge to 300, 500 and 700 respectively, and fill the empty pixels by mirroring the images. 

Traditional cross entropy loss for multi-class classification introduces competition between different classes, thus it is not suitable for images with multiple labels. We compute the loss with binary cross entropy criteria for each class separately, and sum up the error gradients from losses of all classes for back-propagation. 

\subsection{Cascade-style Proposal Verification}
By setting thresholds on proposal scores, a small subset of image boxes which might contain objects are selected. Similar to object detection frameworks, we run CNN classifiers on the selected boxes. The proposal step also serves as a filter whose goal is to preserve the object boxes with high recall rate, while removing the easy negatives. The verification classifiers then address a more focused problem on a smaller set of instances. Connecting the two steps is essentially the same as training a cascade of classifiers.

\textbf{Verification network architecture.} As a later classifier in the cascade, accuracy is more important than speed. We choose the VGG-16 network architecture~\cite{Simonyan14c}. Compared with AlexNet variants, it offers better accuracy for most visual recognition tasks, but is also slower and more memory demanding. We use the VGG-16 model parameters released by the authors, which was trained on 1,000 ImageNet categories. We use the same binary cross entropy criterion to compute losses. To make the training process faster, we only fine-tune the final two fully-connected layers and freeze all previous layers.

\begin{algorithm}
    \SetKwInOut{Input}{Input}
    \SetKwInOut{Output}{Output}
    \Input{Training images with proposal scores $\mathcal{I}_p$, batch size $b$, threshold $t\in [0,1]$}
    \While{stopping criteria not met}
    {
      Randomly select $b$ images $I_1,...,I_b$ from $\mathcal{I}_p$;\\
      Initialize mini-batch $\mathbf{T}$;\\
      \For{$j=1,b$}
      {
         \If{$I_j$ has proposal with score $\ge t$}
         {
         Randomly sample a proposal $\mathbf{l}$ where $P(I_j, c, \mathbf{l})\ge t$;\\
         Set the sample's active class to $c$;\\
         Add proposed region to $\mathbf{T}$;
         }
         \Else
         {
         Resize and add full image to $\mathbf{T}$;\\
         Set all classes as active;
         }
      }
      Forward pass with $\mathbf{T}$;\\
      Compute loss for the active class of each sample;\\
      Update model parameters.\\
    }
\caption{Mini-batch sampling algorithm for training cascade classifier with stochastic gradient descent.}
\label{alg:sample}
\end{algorithm}

\textbf{Training strategy for the cascade.} Ideally, we want the verification network to handle \textit{hard} examples from both positive and negative data. When a proposed region from an image not containing a given label has a high score of that class, we know it is a hard negative. However, it is impossible to tell a hard positive from background without using bounding box annotations. We attempt to avoid using background by selecting only the top scoring image region for each positive class. This results in significant over-fitting and poor generalizability for the trained verification net. 

The main problem with the above sampling strategy is that for positive instances, only \textit{easy} examples which have been learned well are preserved. To fix this, we use a random sampling strategy as described in Algorithm~\ref{alg:sample}. For each image, we randomly select an image box whose proposal score is higher than threshold $t$ for class $c$. In practice, the threshold is set to a relative low value (\eg $0.1$). If $c$ is labeled as positive for the image, we treat the box as a positive instance (though it might belong to background), and otherwise negative. Note that the sampled box could be easy negatives for classes beyond $c$. To avoid oversampling the easy negatives, we set $c$ as the \textit{active class} during back-propagation and only compute the loss for the active class.

\begin{figure}
  \centering
    \includegraphics[width=1.05\linewidth]{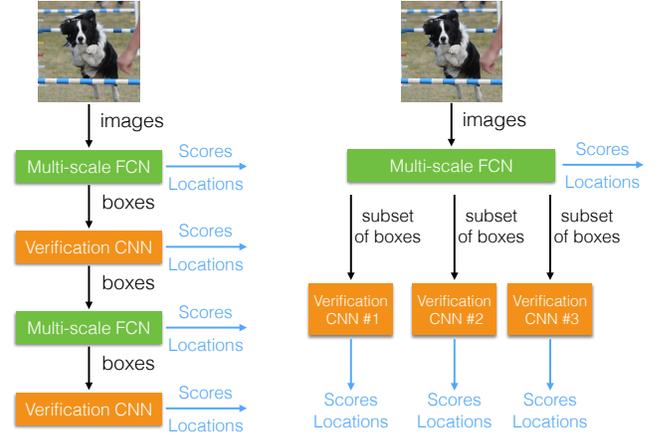}
  \caption{The chain-structure and tree-structure cascades we used in implementing ProNet.}
  \label{fig:chain_tree}
\end{figure}

\begin{figure*}
\centering
\begin{tabular}{ccc}
\includegraphics[width=0.28\linewidth]{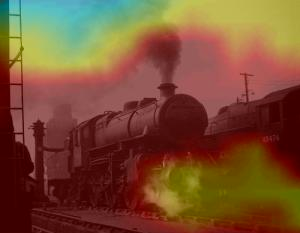}&\includegraphics[width=0.28\linewidth]{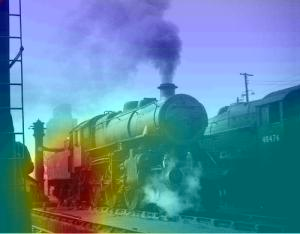}&\includegraphics[width=0.28\linewidth]{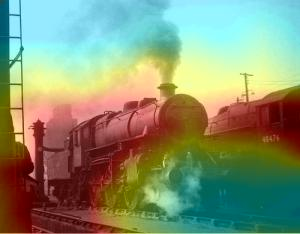}\\
Average pooling & Max pooling & LSE pooling
\end{tabular}  
\caption{Heat map for class \textit{train} generated by proposal network trained with average pooling, max pooling and LSE pooling respectively.}
\label{fig:pool_compare}
\end{figure*}

\textbf{Inference with cascade.} During inference, an image is passed to the proposal generation FCN to compute proposal scores. A small subset of proposed boxes with high scores are then passed to the verification network. For each class, we select the top $k$ scoring proposals if the scores are higher than threshold $t$. We then use the following equation to combine the outputs from both networks:
\begin{equation}
s_c = \left\{
\begin{aligned}[lr]
\max_{l\in\mathcal{L}_c} s^{l}_c&\textrm{\qquad if $\mathcal{L}_c \neq \emptyset$}\\
s^{p}_c&\textrm{\qquad otherwise}
\end{aligned}
\right.
\end{equation}
where $\mathcal{L}_c$ is the set of selected proposals for class $c$, $s^p_c$ is the score of class $c$ from the proposal network after LSE pooling, and $s^l_c$ is the verification network's output for class $c$ on region $l$. When no proposal is selected, we preserve scores from the proposal network without calibration as they are typically low.

\textbf{Discussion.} Decomposing classification into cascade of proposal and verification networks allows the system to achieve high accuracy while maintaining a reasonable computational cost. It is also a flexible framework for different design choices. For example, one could decide to verify a subset of object classes which require higher accuracy. With the cascade training algorithm, we can build tree-structured cascaded neural networks, where each branch focuses on a subset of categories. We can also extend the cascade to have more stages, and train the new stages with newly annotated training data. Figure~\ref{fig:chain_tree} illustrates these structures.

\section{Experiments}
\textbf{Experimental setup.} We work with the PASCAL VOC 2012 dataset~\cite{Everingham10} and the MS COCO dataset~\cite{DBLP:journals/corr/LinMBHPRDZ14}. VOC 2012 has 5,000 images for training, 5,000 for validation and 10,000 for testing. There are 20 object classes in total. COCO has 80,000 images for training and 40,000 images for validation. It has 80 object classes in 12 super-categories.

We evaluated ProNet on object classification and point-based object localization tasks. For object classification, we use the average precision metric. We used VOC's result server to compute \textit{average precisions} on the VOC 2012 dataset. For point-based object localization, we use the criteria introduced in~\cite{Oquab_2015_CVPR}. For every image and every class, we output a location with maximum response for that class. The location is deemed correct if it falls into any bounding box associated with that class, with a tolerance of 18 pixels as used in~\cite{Oquab_2015_CVPR}. This information is then used to compute average precision. Although object extent is not evaluated, the metric remains challenging as shown by~\cite{Oquab_2015_CVPR}. To generate localization coordinates for evaluation, we kept track of the image boxes which give highest responses at each stage, and used the center point of the selected boxes.

We tried different values of hyper-parameter $r$ for LSE pooling, and found that $r\in[5,12]$ generally gave good performance. We fixed $r=10$ in all the following experiments. We used the stochastic gradient descent algorithm for training. To train proposal network, the learning rate was set to 0.01; to train verification network, the learning rate was set to 0.001. We set the  filtering threshold for cascade to 0.1.

\textbf{Which pooling method is better?} We compare \textit{maximum pooling}, \textit{average pooling} and \textit{LSE pooling} methods to train proposal network with image-level supervision. Table~\ref{tab:voc12_val} lists the classification and localization performance of the three different pooling methods. We can see that LSE achieves the best classification mAP. Average pooling is 3.7\% worse than LSE, which we believe is because it assigns equal importance to foreground and background. Max pooling is 1.4\% worse; compared with LSE pooling, it only uses a single patch to generate image-level score, thus is more sensitive to noise and model initialization during training.

\begin{table}
\small
\center
\begin{tabular}{ccc}
\hline
Method & Classification & Localization\\
\hline
\rowcolor{gray!30}Oquab et al.~\cite{Oquab_2015_CVPR} & 81.8 & 74.5\\
RCNN~\cite{DBLP:journals/corr/GirshickDDM13} & 79.2 & 74.8\\
\rowcolor{gray!30}Fast RCNN~\cite{DBLP:journals/corr/Girshick15} & 87.0 & \textbf{81.9}\\
\hline
\hline
Proposal (Max) & 83.4 & 72.5\\
\rowcolor{gray!30}Proposal (Mean) & 81.1 & 62.8\\
Proposal (LSE) & \textbf{84.8} & \textbf{74.8}\\
\hline\hline
\rowcolor{gray!30}Cascade & 88.1 & 77.7\\
Second Cascade & \textbf{89.0} & \textbf{78.5}\\
\hline
\end{tabular}
\caption{Classification and localization mAPs on VOC 2012 validation set.  Higher mAP is better. Second cascade uses additional training data from MS COCO.}
\label{tab:voc12_val}
\end{table}

We also generated visualizations to study the impact of pooling method on trained models. Figure~\ref{fig:pool_compare} shows heat maps of the class \textit{train} when different models are applied to the same image. We can see that the model trained by average pooling has high activations not only on the train but also on part of the background. For max pooling, only the wheel of the train has high response, presumably because it is the most discriminative for the train. Model trained by LSE pooling has high response on the train, but not on the background.

\begin{figure*}[t]
  \centering
    \includegraphics[width=0.9\linewidth]{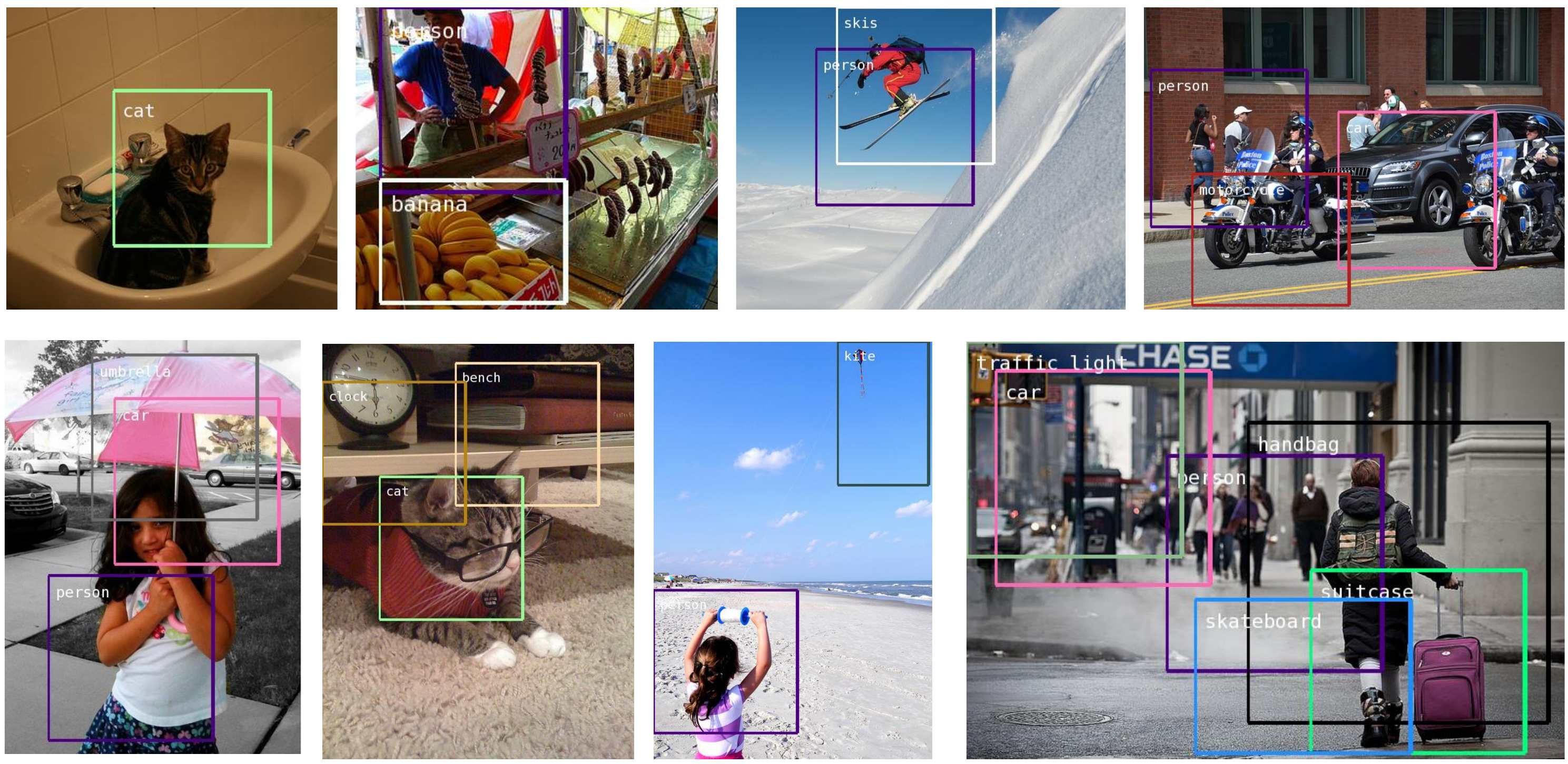}
  \caption{Object localization examples on COCO images. For each object class, we show the box with max score if greater than 0.8. Center point of each box is used for point-based localization evaluation.}
  \label{fig:loc_vis}
\end{figure*}

\textbf{Does cascade help?} We study the impact of adding cascaded classifiers on classification and localization performance. We first use a single level of cascade with one multi-scale FCN and one verification network. For each image and each class, we selected the top 3 regions per scale if their scores are higher than 0.1. The average number of regions to be verified is 24 per image. In Table~\ref{tab:voc12_val}, we can see that on PASCAL VOC 2012, using a cascade helps improve classification mAP by 3.3\% and localization mAP by 2.9\%.

\begin{table}
\small
\center
\begin{tabular}{ccc}
\hline
Method & Classification & Localization\\
\hline
Oquab et al.~\cite{Oquab_2015_CVPR} & 62.8 & 41.2\\
\rowcolor{gray!30}Proposal & 67.8 & 43.5\\
Chain cascade & 69.2 & 45.4\\
\rowcolor{gray!30}Tree cascade & \textbf{70.9} & \textbf{46.4}\\
\hline
\end{tabular}
\caption{Classification and localization mAPs on COCO validation set. Higher mAP is better.}
\label{tab:coco_result}
\end{table}

\begin{table}
\small
\center
\begin{tabular}{cccc}
\hline
$k$ & Ave. \#Proposals & Cls. & Loc.\\
\hline
\rowcolor{gray!30}1 & 9.0 & 87.7 & 76.3\\
2 & 16.6 & 87.9 & 76.8\\
\rowcolor{gray!30}3 & 23.9 & 88.1 & 77.1\\
\hline
\hline
Fast RCNN & \#Proposals & Cls. & Loc.\\
\hline
\rowcolor{gray!30}&10&43.2&34.7\\
&50&70.1&63.2\\
\rowcolor{gray!30}&500&85.8&80.8\\
&1000&87.0&81.9\\
\hline
\end{tabular}
\caption{Impact of number of boxes passed to verification network on VOC 2012 validation set. We also compare the impact of selective search proposal number for fast RCNN.}
\label{tab:proposal_number}
\end{table}


\textbf{Is a longer cascade better?} We are interested in observing how the performance changes with more levels of cascade. For this purpose, we first trained another set of proposal and verification networks using PASCAL VOC data alone, but found that the network overfitted easily. Since the training set of VOC 2012 has only 5,000 images, we found that the first set of proposal and verification networks ``perfectly solved'' this training set, leaving little room to improve its generalizability.

In light of this, we used the 80,000 images from COCO training set as complementary data source. It covers the 20 categories used in VOC but also has 60 other categories. Rather than re-training all the networks by combining VOC and COCO data, we take that the previous CNNs in the cascade have already been trained and fixed, and only train new CNNs with the extra data. Note that our cascade architecture offers a natural way to select the challenging instances from such incoming images.

The final row in Table~\ref{tab:voc12_val} shows the mAPs after adding a new set of cascades trained from COCO images. We can see that it offers another 1\% improvement over the previous cascade, which indicates that it is desirable to train a longer cascade when more training data becomes available.

\textbf{Expanding cascades into trees.} We also investigated the effect of building tree-structured cascades. COCO dataset is used for evaluation as it has 3 times more categories than VOC. 

We trained 12 verification networks corresponding to the 12 super-categories of COCO. Each network focuses on a single super-category, and processes the sampled boxes whose active classes belong to that super-category. At test time, each proposed box only goes through a single root to leaf path in the tree. The final row of Table~\ref{tab:coco_result} shows its classification and localization performance. We can see that compared with the chain structured cascade, tree-structured cascade achieves better performance, probably because it trains the neural networks to be focused on a small subset of similar categories.

\begin{table*}
\footnotesize
\tabcolsep=0.06cm
\centering
\begin{tabular}{ccccccccccccccccccccccc}    
\hline
Method & BBox  & plane & bike & bird & boat & btl & bus & car & cat & chair & cow & tabl & dog & hors & moto & pers & plant & sheep & sofa & train & tv & mAP\\
\hline
NUS-PSL~\cite{NUS-PSL} & Yes & 97.3 & 84.2 & 80.8 & 85.3 & 60.8 & 89.9 & 86.8 & 89.3 & 75.4 & 77.8 & 75.1 & 83.0 & 87.5 & 90.1 & 95.0 & 57.8 & 79.2 & 73.4 & 94.5 & 80.7 & 82.2\\
\rowcolor{gray!30} Oquab et al.~\cite{DBLP:conf/cvpr/OquabBLS14} & Yes & 94.6 & 82.9 & 88.2 & 84.1 & 60.3 & 89.0 & 84.4 & 90.7 & 72.1 & 86.8 & 69.0 & 92.1 & 93.4 & 88.6 & \textbf{96.1} & 64.3 & 86.6 & 62.3 & 91.1 & 79.8 & 82.8\\
NUS~\cite{DBLP:journals/corr/WeiXHNDZY14}+\cite{NUS-PSL}$^\star$ & Yes & \textbf{98.9} & \textbf{91.8} & \textbf{94.8} & \textbf{92.4} & \textbf{72.6} & \textbf{95.0} & \textbf{91.8} & \textbf{97.4} & \textbf{85.2} & \textbf{92.9} & \textbf{83.1} & \textbf{96.0} & \textbf{96.6} & \textbf{96.1} & 94.9 & \textbf{68.4} & \textbf{92.0} & \textbf{79.6} & \textbf{97.3} & \textbf{88.5} & \textbf{90.3}\\
\hline
\hline
\rowcolor{gray!30}Zeiler et al.~\cite{DBLP:journals/corr/ZeilerF13} & No & 96.0 & 77.1 & 88.4 & 85.5 & 55.8 & 85.8 & 78.6 & 91.2 & 65.0 & 74.4 & 67.7 & 87.8 & 86.0 & 85.1 & 90.9 & 52.2 & 83.6 & 61.1 & 91.8 & 76.1 & 79.0\\
Chatfield et al.~\cite{DBLP:journals/corr/ChatfieldSVZ14} & No & 96.8 & 82.5 & 91.5 & 88.1 & 62.1 & 88.3 & 81.9 & 94.8 & 70.3 & 80.2 & 76.2 & 92.9 & 90.3 & 89.3 & 95.2 & 57.4 & 83.6 & 66.4 & 93.5 & 81.9 & 83.2\\
\rowcolor{gray!30}NUS-HCP~\cite{DBLP:journals/corr/WeiXHNDZY14} & No & 97.5 & 84.3 & 93.0 & 89.4 & 62.5 & 90.2 & 84.6 & 94.8 & 69.7 & 90.2 & 74.1 & 93.4 & 93.7 & 88.8 & 93.2 & 59.7 & 90.3 & 61.8 & 94.4 & 78.0 & 84.2 \\
Oquab et al.~\cite{Oquab_2015_CVPR} & No & 96.7 & 88.8 & 92.0 & 87.4 & 64.7 & 91.1 & 87.4 & 94.4 & 74.9 & 89.2 & 76.3 & 93.7 & 95.2 & 91.1 & 97.6 & 66.2 & 91.2 & 70.0 & 94.5 & 83.7 & 86.3\\
\rowcolor{gray!30}Simonyan et al.~\cite{Simonyan14c}$^\star$ & No & \textbf{99.0} & 88.8 & \textbf{95.9} & \textbf{93.8} & 73.1 & 92.1 & 85.1 & \textbf{97.8} & \textbf{79.5} & \textbf{91.1} & \textbf{83.3} & \textbf{97.2} & \textbf{96.3} & 94.5 & 96.9 & 63.1 & \textbf{93.4} & 75.0 & \textbf{97.1} & 87.1& 89.0\\
Our Proposal & No & 97.0 & 88.3 & 92.4 & 89.8 & 67.9 & 90.7 & 86.2 & 95.5 & 73.0 & 85.5 & 76.7 & 94.8 & 91.1 & 91.9 & 97.0 & 66.1 & 87.8 & 68.1 & 94.1 & 87.0 & 86.0\\
\rowcolor{gray!30}Our Cascade$^\star$ & No & 97.6 & \textbf{91.3} & 94.3 & 93.2 & \textbf{74.3} & \textbf{93.0} & \textbf{88.5} & 96.8 & 78.4 & 90.7 & 80.1 & 96.3 & 95.2 & \textbf{94.8} & \textbf{98.0} & \textbf{70.9} & 90.3 & \textbf{75.8} & 96.3 & \textbf{89.4} & \textbf{89.3}\\
\hline
\end{tabular}
\normalsize
\caption{Classification performance measured by average precision on PASCAL VOC 2012 test set. BBox column indicates whether the training algorithm uses bounding box annotation or not. $\star$: uses VGG-16 models.}
\label{voc12_cls}
\end{table*}

\begin{table*}
\footnotesize
\tabcolsep=0.06cm
\centering
\begin{tabular}{ccccccccccccccccccccccc}    
\hline
Method & BBox & plane & bike & bird & boat & btl & bus & car & cat & chair & cow & tabl & dog & hors & moto & pers & plant & sheep & sofa & train & tv & mAP\\
\hline
RCNN~\cite{DBLP:journals/corr/GirshickDDM13} & Yes & 92.0 & 80.8 & 80.8 & 73.0 & \textbf{49.9} & 86.8 & 77.7 & 87.6 & 50.4 & 72.1 & 57.6 & 82.9 & 79.1 & 89.8 & 88.1 & 56.1 & 83.5 & 50.1 & 81.5 & 76.6 & 74.8\\
\rowcolor{gray!30}Fast RCNN~\cite{DBLP:journals/corr/Girshick15} & Yes & \textbf{95.2} & \textbf{88.2} & \textbf{88.4} & \textbf{77.9} & 49.0 & \textbf{93.4} & \textbf{83.6} & \textbf{95.1} & \textbf{59.4} & \textbf{86.6} & \textbf{71.0} & \textbf{92.6} & \textbf{93.1} & \textbf{93.0} & \textbf{92.2} & \textbf{58.2} & \textbf{88.0} & \textbf{63.6} & \textbf{91.9} & \textbf{77.3} & \textbf{81.9}\\
\hline\hline
Oquab et al.~\cite{Oquab_2015_CVPR} & No & 90.3 & 77.4 & 81.4 & 79.2 & 41.4 & 87.8 & 66.4 & 91.0 & 47.3 & \textbf{83.7} & 55.1 & 88.8 & \textbf{93.6} & 85.2 & \textbf{87.4} & 43.5 & 86.2 & 50.8 & 86.8 & 66.5 & 74.5\\
\rowcolor{gray!30}Our Proposal & No & 91.6 & 82.0 & 85.1 & 78.6 & 45.9 & 87.9 & 67.1 & 92.2 & 51.0 & 72.9 & 60.8 & 89.3 & 85.1 & 85.3 & 86.4 & 45.6 & 83.5 & 55.1 & 85.6 & 65.9 & 74.8\\
Our Cascade & No & \textbf{92.6} & \textbf{85.6} & \textbf{87.4} & \textbf{79.6} & \textbf{48.3} & \textbf{88.7} & \textbf{68.9} & \textbf{94.2} & \textbf{54.6} & 83.2 & \textbf{62.8} & \textbf{92.0} & 89.9 & \textbf{88.2} & 87.1 & \textbf{49.2} & \textbf{86.9} & \textbf{57.2} & \textbf{86.8} & \textbf{70.0} & \textbf{77.7}\\
\hline
\end{tabular}
\normalsize
\caption{Localization performance measured by average precision on PASCAL VOC 2012 validation set.}
\label{voc12_loc}
\end{table*}


\textbf{Comparison with detection based approaches.} We compare our proposed framework with two recent state-of-the-art object detection methods: RCNN~\cite{DBLP:journals/corr/GirshickDDM13} and Fast RCNN~\cite{DBLP:journals/corr/Girshick15}. Unlike our framework, they require bounding box annotations for training. Both methods use \textit{selective search} to generate object proposals and CNNs for classification. RCNN uses AlexNet~\cite{NIPS2012_4824} pre-trained from ImageNet, while fast RCNN uses VGG-16~\cite{Simonyan14c} pre-trained from ImageNet. To generate classification and localization results, for each class we select the detection output with maximum confidence score, and use the center of the detected bounding box for localization evaluation.

We first fix the number of window proposals to 1000 for RCNN and fast RCNN. Table~\ref{tab:voc12_val} shows the performance comparison. We can see that for classification, our proposed framework outperforms both RCNN and fast RCNN. For localization, our proposed framework outperforms RCNN, but is 4\% worse than fast RCNN. 

We also study the impact of number of proposed boxes on our system's performance. For this purpose, we let the proposal network select top $k=1,2,3$ regions per scale for each class, and compute the average number of proposed boxes per image. For comparison, we ask fast RCNN to use up to 10, 50, 500 and 1000 selective search proposals per image. Table~\ref{tab:proposal_number} shows the classification and localization performances respectively. We can see that ProNet is quite robust to the number of proposed boxes, and achieves reasonably good performance with only 9 boxes on average. This confirms that ProNet offers better accuracy with relatively small computational overhead. Meanwhile, fast RCNN requires many more proposals to reach peak performance, presumably because the selective search proposals are for general \textit{objectness} and not optimized for object classification in cascade fashion.

\textbf{Comparison with other weakly-supervised methods.} We compare ProNet with several state-of-the-art object classification frameworks. Classification and localization performance on PASCAL VOC 2012 are shown in Table~\ref{voc12_cls} and Table~\ref{voc12_loc} respectively. Table~\ref{tab:coco_result} and Figure~\ref{fig:loc_vis} show results and localization examples on COCO dataset. Among the compared systems, Oquab et al. and NUS-HCP use CNNs pre-trained on the expanded ImageNet data with more than 1500 categories, which has been shown to be useful for classification. Since ProNet uses cascades or trees of CNNs, it can apply a more powerful CNN model VGG-16 with small computational overhead. This helps our system outperform most of the previous state-of-the-art systems significantly on both datasets. ProNet is also slightly better than Simonyan et al. which extracts VGG-16 features at three different scales over full images. Their system is 3x to 6x slower than our cascade at test time.


\textbf{Limitation.} We evaluate ProNet using the standard IOU metric, which considers object extent as well as location. Since the boxes generated by our proposal CNN have fixed aspect ratios, we follow~\cite{Oquab_2015_CVPR} to aggregate the heat maps over 1000 bounding box proposals generated by selective search per image. No bounding box regression is conducted. Cascade CNN is then used to verify the high-scoring proposals. On PASCAL VOC 2012 validation set, our proposal CNN has an mAP of 13.0\% when overlap threshold is 0.5. The cascade CNN improves the mAP to 15.5\%. Although both results are higher than 11.7\% as reported by~\cite{Oquab_2015_CVPR}, there is still a huge gap between the state-of-the-art object detection pipelines. Our proposal network tends to select the most discriminative / confusing parts of objects, which is good for cascade classification but bad for getting full object extents. Separating and counting multiple objects are also challenging issues.


\section{Conclusion}
We proposed ProNet, a cascaded neural network for object classification and localization. ProNet learns to propose object-specific boxes by multi-scale FCNs trained from image-level annotations. It then sends a small subset of promising boxes to latter CNNs for verification. Detailed experimental evaluations have shown the effectiveness of ProNet on the challenging PASCAL VOC 2012 dataset and MS COCO dataset.

{\small
\quad\newline
\textbf{Acknowledgement:}
We would like to thank Sergey Zagoruyko for help with fast RCNN experiments;
Pedro O. Pinheiro, Bolei Zhou, Maxime Oquab, Jo{\"{e}}l Legrand, Yuandong Tian, L{\'{e}}on Bottou and Florent Perronnin for valuable discussions.
}

{\small
\bibliographystyle{ieee}
\bibliography{egbib}

\begin{thebibliography}{10}\itemsep=-1pt

\bibitem{Bilen_2015_CVPR}
H.~Bilen, M.~Pedersoli, and T.~Tuytelaars.
\newblock Weakly supervised object detection with convex clustering.
\newblock In {\em CVPR}, 2015.

\bibitem{DBLP:journals/corr/ChatfieldSVZ14}
K.~Chatfield, K.~Simonyan, A.~Vedaldi, and A.~Zisserman.
\newblock Return of the devil in the details: Delving deep into convolutional
  nets.
\newblock In {\em BMVC}, 2014.

\bibitem{chen2013neil}
X.~Chen, A.~Shrivastava, and A.~Gupta.
\newblock {NEIL}: Extracting visual knowledge from web data.
\newblock In {\em ICCV}, 2013.

\bibitem{DBLP:conf/cvpr/CinbisVS14}
R.~G. Cinbis, J.~J. Verbeek, and C.~Schmid.
\newblock Multi-fold {MIL} training for weakly supervised object localization.
\newblock In {\em CVPR}, 2014.

\bibitem{DBLP:journals/corr/DaiH015}
J.~Dai, K.~He, and J.~Sun.
\newblock Boxsup: Exploiting bounding boxes to supervise convolutional networks
  for semantic segmentation.
\newblock In {\em ICCV}, 2015.

\bibitem{DBLP:conf/cvpr/DivvalaFG14}
S.~K. Divvala, A.~Farhadi, and C.~Guestrin.
\newblock Learning everything about anything: Webly-supervised visual concept
  learning.
\newblock In {\em CVPR}, 2014.

\bibitem{Everingham10}
M.~Everingham, L.~Van~Gool, C.~K.~I. Williams, J.~Winn, and A.~Zisserman.
\newblock The pascal visual object classes (voc) challenge.
\newblock {\em IJCV}, 2010.

\bibitem{DBLP:journals/corr/Girshick15}
R.~B. Girshick.
\newblock Fast {R-CNN}.
\newblock In {\em ICCV}, 2015.

\bibitem{DBLP:journals/corr/GirshickDDM13}
R.~B. Girshick, J.~Donahue, T.~Darrell, and J.~Malik.
\newblock Rich feature hierarchies for accurate object detection and semantic
  segmentation.
\newblock In {\em CVPR}, 2014.

\bibitem{DBLP:journals/corr/GongWGL14}
Y.~Gong, L.~Wang, R.~Guo, and S.~Lazebnik.
\newblock Multi-scale orderless pooling of deep convolutional activation
  features.
\newblock In {\em ECCV}, 2014.

\bibitem{DBLP:conf/iccv/GraumanD05}
K.~Grauman and T.~Darrell.
\newblock The pyramid match kernel: Discriminative classification with sets of
  image features.
\newblock In {\em ICCV}, 2005.

\bibitem{NIPS2012_4824}
A.~Krizhevsky, I.~Sutskever, and G.~E. Hinton.
\newblock Imagenet classification with deep convolutional neural networks.
\newblock In {\em NIPS}. 2012.

\bibitem{DBLP:journals/cviu/Fei-FeiFP07}
F.~Li, R.~Fergus, and P.~Perona.
\newblock Learning generative visual models from few training examples: An
  incremental bayesian approach tested on 101 object categories.
\newblock {\em CVIU}, 2007.

\bibitem{DBLP:conf/cvpr/LiLSBH15}
H.~Li, Z.~Lin, X.~Shen, J.~Brandt, and G.~Hua.
\newblock A convolutional neural network cascade for face detection.
\newblock In {\em CVPR}, 2015.

\bibitem{DBLP:journals/corr/LinMBHPRDZ14}
T.~Lin, M.~Maire, S.~Belongie, L.~D. Bourdev, R.~B. Girshick, J.~Hays,
  P.~Perona, D.~Ramanan, P.~Doll{\'{a}}r, and C.~L. Zitnick.
\newblock Microsoft {COCO:} common objects in context.
\newblock {\em CoRR}, abs/1405.0312, 2014.

\bibitem{DBLP:journals/corr/LongSD14}
J.~Long, E.~Shelhamer, and T.~Darrell.
\newblock Fully convolutional networks for semantic segmentation.
\newblock In {\em CVPR}, 2015.

\bibitem{DBLP:journals/ijcv/Lowe04}
D.~G. Lowe.
\newblock Distinctive image features from scale-invariant keypoints.
\newblock {\em IJCV}, 2004.

\bibitem{DBLP:conf/cvpr/OquabBLS14}
M.~Oquab, L.~Bottou, I.~Laptev, and J.~Sivic.
\newblock Learning and transferring mid-level image representations using
  convolutional neural networks.
\newblock In {\em CVPR}, 2014.

\bibitem{Oquab_2015_CVPR}
M.~Oquab, L.~Bottou, I.~Laptev, and J.~Sivic.
\newblock Is object localization for free? - weakly-supervised learning with
  convolutional neural networks.
\newblock In {\em CVPR}, 2015.

\bibitem{DBLP:journals/corr/PapandreouCMY15}
G.~Papandreou, L.~Chen, K.~Murphy, and A.~L. Yuille.
\newblock Weakly- and semi-supervised learning of a {DCNN} for semantic image
  segmentation.
\newblock In {\em ICCV}, 2015.

\bibitem{DBLP:journals/corr/PathakSLD14}
D.~Pathak, E.~Shelhamer, J.~Long, and T.~Darrell.
\newblock Fully convolutional multi-class multiple instance learning.
\newblock In {\em ICLR}, 2015.

\bibitem{Perronnin07FV}
F.~Perronnin and C.~Dance.
\newblock Fisher kernels on visual vocabularies for image categorization.
\newblock In {\em CVPR}, 2007.

\bibitem{pinheiro:2015a}
P.~H.~O. Pinheiro and R.~Collobert.
\newblock From image-level to pixel-level labeling with convolutional networks.
\newblock In {\em CVPR}, 2015.

\bibitem{ILSVRCarxiv14}
O.~Russakovsky, J.~Deng, H.~Su, J.~Krause, S.~Satheesh, S.~Ma, Z.~Huang,
  A.~Karpathy, A.~Khosla, M.~Bernstein, A.~C. Berg, and L.~Fei-Fei.
\newblock {ImageNet Large Scale Visual Recognition Challenge}, 2014.

\bibitem{DBLP:conf/eccv/RussakovskyLYF12}
O.~Russakovsky, Y.~Lin, K.~Yu, and F.~Li.
\newblock Object-centric spatial pooling for image classification.
\newblock In {\em ECCV}, 2012.

\bibitem{DBLP:journals/ijcv/SanchezPMV13}
J.~S{\'{a}}nchez, F.~Perronnin, T.~Mensink, and J.~J. Verbeek.
\newblock Image classification with the fisher vector: Theory and practice.
\newblock {\em IJCV}, 2013.

\bibitem{DBLP:journals/corr/SermanetEZMFL13}
P.~Sermanet, D.~Eigen, X.~Zhang, M.~Mathieu, R.~Fergus, and Y.~LeCun.
\newblock Overfeat: Integrated recognition, localization and detection using
  convolutional networks.
\newblock {\em CoRR}, abs/1312.6229, 2013.

\bibitem{Simonyan14c}
K.~Simonyan and A.~Zisserman.
\newblock Very deep convolutional networks for large-scale image recognition.
\newblock In {\em ICLR}, 2015.

\bibitem{DBLP:conf/icml/SongGJMHD14}
H.~O. Song, R.~B. Girshick, S.~Jegelka, J.~Mairal, Z.~Harchaoui, and
  T.~Darrell.
\newblock On learning to localize objects with minimal supervision.
\newblock In {\em ICML}, 2014.

\bibitem{DBLP:conf/cvpr/SunWT13}
Y.~Sun, X.~Wang, and X.~Tang.
\newblock Deep convolutional network cascade for facial point detection.
\newblock In {\em CVPR}, 2013.

\bibitem{DBLP:journals/corr/ToshevS13}
A.~Toshev and C.~Szegedy.
\newblock Deeppose: Human pose estimation via deep neural networks.
\newblock In {\em CVPR}, 2014.

\bibitem{DBLP:journals/ijcv/UijlingsSGS13}
J.~R.~R. Uijlings, K.~E.~A. van~de Sande, T.~Gevers, and A.~W.~M. Smeulders.
\newblock Selective search for object recognition.
\newblock {\em IJCV}, 2013.

\bibitem{DBLP:conf/cvpr/ViolaJ01}
P.~A. Viola and M.~J. Jones.
\newblock Rapid object detection using a boosted cascade of simple features.
\newblock In {\em CVPR}, 2001.

\bibitem{DBLP:conf/cvpr/WangYYLHG10}
J.~Wang, J.~Yang, K.~Yu, F.~Lv, T.~S. Huang, and Y.~Gong.
\newblock Locality-constrained linear coding for image classification.
\newblock In {\em CVPR}, 2010.

\bibitem{DBLP:journals/corr/WeiXHNDZY14}
Y.~Wei, W.~Xia, J.~Huang, B.~Ni, J.~Dong, Y.~Zhao, and S.~Yan.
\newblock {CNN:} single-label to multi-label.
\newblock {\em CoRR}, abs/1406.5726, 2014.

\bibitem{wu2015webconcept}
J.~Wu, Y.~Yu, C.~Huang, and K.~Yu.
\newblock Deep multiple instance learning for image classification and
  auto-annotation.
\newblock {\em CVPR}, 2015.

\bibitem{NUS-PSL}
S.~Yan, J.~Dong, Q.~Chen, Z.~Song, Y.~Pan, W.~Xia, H.~Zhongyang, Y.~Hua, and
  S.~Shen.
\newblock Generalized hierarchical matching for subcategory aware object
  classification.
\newblock In {\em ECCV Workshop}, 2012.

\bibitem{DBLP:journals/corr/ZeilerF13}
M.~D. Zeiler and R.~Fergus.
\newblock Visualizing and understanding convolutional networks.
\newblock In {\em ECCV}, 2014.

\bibitem{zhou2015cnnlocalization}
B.~Zhou, A.~Khosla, L.~A., A.~Oliva, and A.~Torralba.
\newblock {Learning Deep Features for Discriminative Localization.}
\newblock {\em CVPR}, 2016.

\bibitem{DBLP:conf/eccv/ZitnickD14}
C.~L. Zitnick and P.~Doll{\'{a}}r.
\newblock Edge boxes: Locating object proposals from edges.
\newblock In {\em ECCV}, 2014.

\end{thebibliography}
}

\end{document}